\documentclass[letterpaper, 10 pt, conference]{ieeeconf}  

\IEEEoverridecommandlockouts  

\usepackage{tikz}
\usepackage{textcomp}
\usepackage{hyperref}
\usepackage{lipsum}

\newcommand\copyrighttext{%
  \footnotesize \textcopyright 2012 IEEE. Personal use of this material is permitted.
  Permission from IEEE must be obtained for all other uses, in any current or future
  media, including reprinting/republishing this material for advertising or promotional
  purposes, creating new collective works, for resale or redistribution to servers or
  lists, or reuse of any copyrighted component of this work in other works.
  DOI: \href{<http://tex.stackexchange.com>}{10.1109/SIMPAR.2018.8376271}}
\newcommand\copyrightnotice{%
\begin{tikzpicture}[remember picture,overlay]
\node[anchor=south,yshift=10pt] at (current page.south) {\fbox{\parbox{\dimexpr\textwidth-\fboxsep-\fboxrule\relax}{\copyrighttext}}};
\end{tikzpicture}%
}

\usepackage{graphicx} 
\usepackage{mathptmx} 
\usepackage{times} 
\usepackage{amsmath} 
\usepackage{amssymb}  
\usepackage{siunitx}
\usepackage{textcomp}
\usepackage[english]{babel}
\usepackage{siunitx} 
\usepackage{subcaption}
\usepackage[font=small,labelfont=bf]{caption}
\usepackage[backend=biber,      
    bibstyle=ieee,              
    citestyle=numeric-comp,     
    sortcites=true,   
    giveninits=true,            
    maxbibnames=3
]{biblatex}

\title{\LARGE \bf
Learning from Outside the Viability Kernel: \\Why we Should Build Robots that can Fall With Grace
}

\author{Steve Heim and Alexander Spr\"{o}witz
\thanks{Dynamic Locomotion Group, Max Planck Institute for Intelligent Systems, Stuttgart, DE
        {\tt\small \{heim, sprowitz\}@.is.mpg.de}
}}

\begin{document}

\maketitle
\copyrightnotice 
\thispagestyle{empty}
\pagestyle{empty}

\begin{abstract}

Despite impressive results using reinforcement learning to solve complex problems from scratch, in robotics this has still been largely limited to model-based learning with very informative reward functions. One of the major challenges is that the reward landscape often has large patches with no gradient, making it difficult to sample gradients effectively. We show here that the robot state-initialization can have a more important effect on the reward landscape than is generally expected. In particular, we show the counter-intuitive benefit of including initializations that are \emph{unviable}, in other words initializing in states that are doomed to fail.
\end{abstract}
\section{INTRODUCTION}

Recent advances in reinforcement learning (RL) have shown a lot of promise, especially with the capability to learn complex policies from scratch, model-free and from very generic reward signals \cite{silver2017mastering}\cite{sutton1998reinforcement}. However it has been difficult to transfer these results into learning directly in robot hardware. Most RL in robotics is model-based and uses highly informative reward functions \cite{kober2013reinforcement}. \par
One of the main challenges is that much of the reward landscape (sometimes called a cost landscape) that a learning algorithm needs to traverse tends to be flat, providing no informative gradient. In classical RL, the ability to run a huge number of trials, often in parallel, with completely random initial conditions for both state and policy parameters helps alleviate the problem. With a robot, this can be prohibitively expensive in terms of time, hardware costs and computation. It is typically necessary to invest substantial effort into the design of a reasonable initial controller and parameterization, and subsequently improve it via RL \cite{heijmink2017learning}\cite{kohl2004policy}, instead of learning policies from scratch. \par
Another common approach is to rely on model-based control methods, and use RL to learn the model, instead of learning the policy \cite{kober2013reinforcement}\cite{levine2015learning}. \par
It is also important to choose an appropriate exploration strategy, and the subfield of intrinsic motivation in particular addresses the problem of getting stuck in a flat portion of the reward landscape \cite{chentanez2005intrinsically}. Alternatively, the reward landscape can be shaped by providing more informative reward functions \cite{abbeel2004apprenticeship}\cite{kalakrishnan2013learning}, or a progressive set of reward functions or environments \cite{bengio2009curriculum}\cite{karpathy2012curriculum}\cite{randlov1998learning}\cite{heim2017shaping}. \par
While these approaches all have merit, we show that the choice of state initialization can have a greater effect on the reward landscape than is usually assumed. The current practice is to initialize from a stable state, as states that fail with a high probability can be crippling. Quick failures often result in no reward and therefore no gradient to learn from, and often damage the robot hardware itself. At the most restrictive, this means the robot always starts within the basin of attraction of a stable controller. In order to accelerate learning, it is important for the learning agent to explore outside this region, and indeed considerable effort has been made to be able to step outside the basin of attraction and into its superset \cite{prajna2004safety}\cite{smitsafe}, the \emph{viability kernel}. \par
A viability kernel \cite{aubin2009viability} is the set of all states from which there is at least one time-evolution that remains confined to a desired region. Sampling states outside the current policy's basin of attraction allows the agent to learn more aggressively and still avoid potentially disastrous failures. When learning a model of the dynamics, the agent can clearly benefit by being even more aggressive and initially visit failure states. These states will often be in areas of the state-space that are otherwise not visited, and data-points sampled there can give additional information to fit a model more accurately. When directly learning a policy in a model-free approach however, the benefit of visiting states from which no stabilizing policy exists is less obvious. \par
The main contribution of this work is to show that it can be beneficial for model-free learning to initialize the robot from states \emph{outside the viability kernel}.

\section{VIABILITY KERNEL OF A RUNNING MODEL}

\subsection{Model}
Our work revolves around the spring-loaded inverted pendulum (SLIP) model, a hybrid dynamic model ubiquitous in both biomechanics and the robotic legged-locomotion community for modeling running or hopping gaits. This model can be fit very accurately to experimental data of many different running animals \cite{blickhan1989spring}\cite{full1999templates}, allows accurate prediction \cite{maus2015constructing}, and also has been used to design controllers for simulations \cite{cnops2015basin}\cite{piovan2015reachability}\cite{uyanik2011adaptive}\cite{wu20133} as well as actual robots \cite{martin2017experimental}\cite{rezazadeh2015spring}\cite{terry2016towards}. Indeed, there has been a lot of effort to give legged robots SLIP-like behavior, either through mechanical design \cite{hubicki2016atrias}\cite{lakatos2017eigenmodes} or control \cite{martin2017experimental}\cite{hutter2010slip}. The two dimensional view of the model, in the sagittal plane parallel to the direction of travel, represents a submanifold of the 3D-space, and the results can be extended to 3D motion both in simulation \cite{wu20133} and hardware \cite{raibert1986legged}. \par
We consider this an ideal model that is both low-dimensional enough to clearly illustrate our result, while also informative and applicable to real-world systems. 
\begin{figure}[tbh]
    \centering
    \includegraphics[width=1\columnwidth]{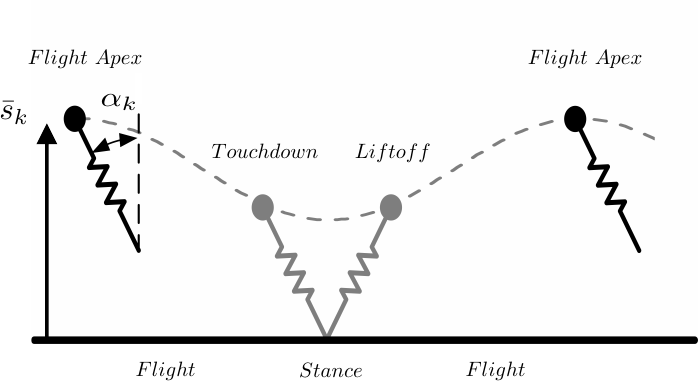}
    \caption{The classical spring-loaded inverted pendulum is characterized by hybrid dynamics: the governing equations of motion switch between flight and stance phases at the touchdown and liftoff events. We have highlighted here the normalized height at apex $\bar{s}_k$ as our one-dimensional state and the landing angle of attack $\alpha_k$ as our one-dimensional action.} 
    \label{fig:slip}
\end{figure}
The model, shown in Fig. \ref{fig:slip}, has a point mass representing the body center of gravity, and a massless spring representing the leg. Touchdown and liftoff conditions govern the switch between flight and stance phases. The model uses a no-slip assumption for the foot during stance phase. At liftoff, the massless leg is instantaneously reset to the landing angle of attack $\alpha_k$. The continuous dynamics are as follow:
\begin{gather*}
\text{Continuous Dynamics} \nonumber \\[5mm]
\text{Flight Dynamics} \nonumber \\
\left[
\begin{array}{c}
\ddot{x}\\
\ddot{y}\\
\end{array}
\right] = \left[
\begin{array}{c}
0\\
-g\\
\end{array}
\right] \\[5mm]
\text{Stance Dynamics} \nonumber \\
\left[
\begin{array}{c}
\ddot{x}\\
\ddot{y}\\
\end{array}
\right] = \frac{k}{m}(l_0 - l)\left[
\begin{array}{c}
\sin(\alpha)\\
\cos(\alpha)\\
\end{array}
\right] - 
\left[
\begin{array}{c}
0 \\
g\\
\end{array}
\right]\\[5mm]
\text{Touchdown Condition}\nonumber \\
y = l_0\cos(\alpha_k) \\[5mm]
\text{Liftoff Condition}\nonumber \\
l = l_0 \\[5mm]
\alpha = \arctan(y/x) - \frac{\pi}{2} \nonumber \\
l = \sqrt{((x-x_f)^2 + y^2)} \nonumber
\end{gather*}

\begin{equation*}
\begin{aligned}[c]
x &: \text{horizontal position}\nonumber\\
y &: \text{vertical position}\nonumber \\
\alpha &: \text{angle of attack}\nonumber \\
x_f &: \text{foot position}\nonumber \\
l &: \text{leg length}\nonumber \\
\end{aligned}
\quad
\begin{aligned}[c]
l_0 &: \text{leg resting length = 1 [$m$]}\nonumber \\
k &: \text{spring stiffness = 8200 [$N/m$]}\nonumber\\ 
\alpha_k &: \text{landing angle of attack [$rad$]} \\
m &: \text{body mass = 80 [$kg$]}\nonumber\\
g &: \text{gravity constant = 9.81 [$m/s^2$]}\nonumber\\ 
\end{aligned}
\end{equation*}
The parameters used are parameters commonly used to model human running \cite{seyfarth2002movement}. \par
As is commonly done \cite{cnops2015basin}\cite{piovan2015reachability}\cite{rummel2008stable}, we use Poincar\'{e} return maps to study the system \cite{westervelt2007feedback}. We numerically integrate the continuous-time dynamics from one apex height to the next, and examine the state at these apex events. This gives us the following discrete mapping:
\begin{gather*}
\text{Discrete Dynamics} \nonumber \\
\text{\small{Continuous Dynamics Integrated from Apex to Apex}} \nonumber \\[5mm]
\text{State at Apex} \nonumber \\
s_k = \left[
\begin{array}{c}
y_{apex}\\
\dot{x}_{apex}\\
\end{array}
\right] \\[5mm]
\text{Normalized Height at Apex} \nonumber \\
\bar{s}_k = \frac{E_g}{E_g + E_k} = \frac{g\ y_{apex}}{\frac{\dot{x}^2_{apex}}{2} + gy_{apex}} \\[5mm]
\text{Apex Transition Dynamics} \nonumber \\
\bar{s}_{k+1} = P(\bar{s}_k, \alpha_k)
\end{gather*}
\begin{align*}
E_g &: \text{Potential Energy} \\
E_k &: \text{Kinetic Energy} \\
P &: \text{Poincar\'{e} map}
\end{align*}\label{eq:energies}
We drop the $x$ coordinate as it is not periodic, and at apex the $\dot{y}$ coordinate is zero by definition, which leaves us with only two coordinates, $y$ and $\dot{x}$. Finally, we exploit the fact that the system is energy-conservative and use the constant total energy constraint to reduce the system to a single state, a normalized hopping height. This results in the map $\bar{s}_{k+1} = P(s_k, \alpha_k)$. Note that we include the landing angle of attack parameter $\alpha_{k}$ as a control input which can be chosen freely at each apex. The original passive model was used to analyze steady-state locomotion and set $\alpha_k$ as a constant parameter; for non-steady-state locomotion, the landing angle of attack is a well studied choice for a control input \cite{wu20133}\cite{raibert1986legged}\cite{palmer2014periodic}, alongside spring-stiffness. \par
\subsection{Transition Matrix and Viability Kernel}
The discrete dynamics of the system have two possible transitions: either a mapping from one apex height to the next apex height, or from an apex height to a failure state, in which the point-mass body hits the ground with $\bar{s}_{k+1} = 0$. \par
\begin{figure*}[hbtp]
    \centering
    \includegraphics[width=0.95\linewidth]{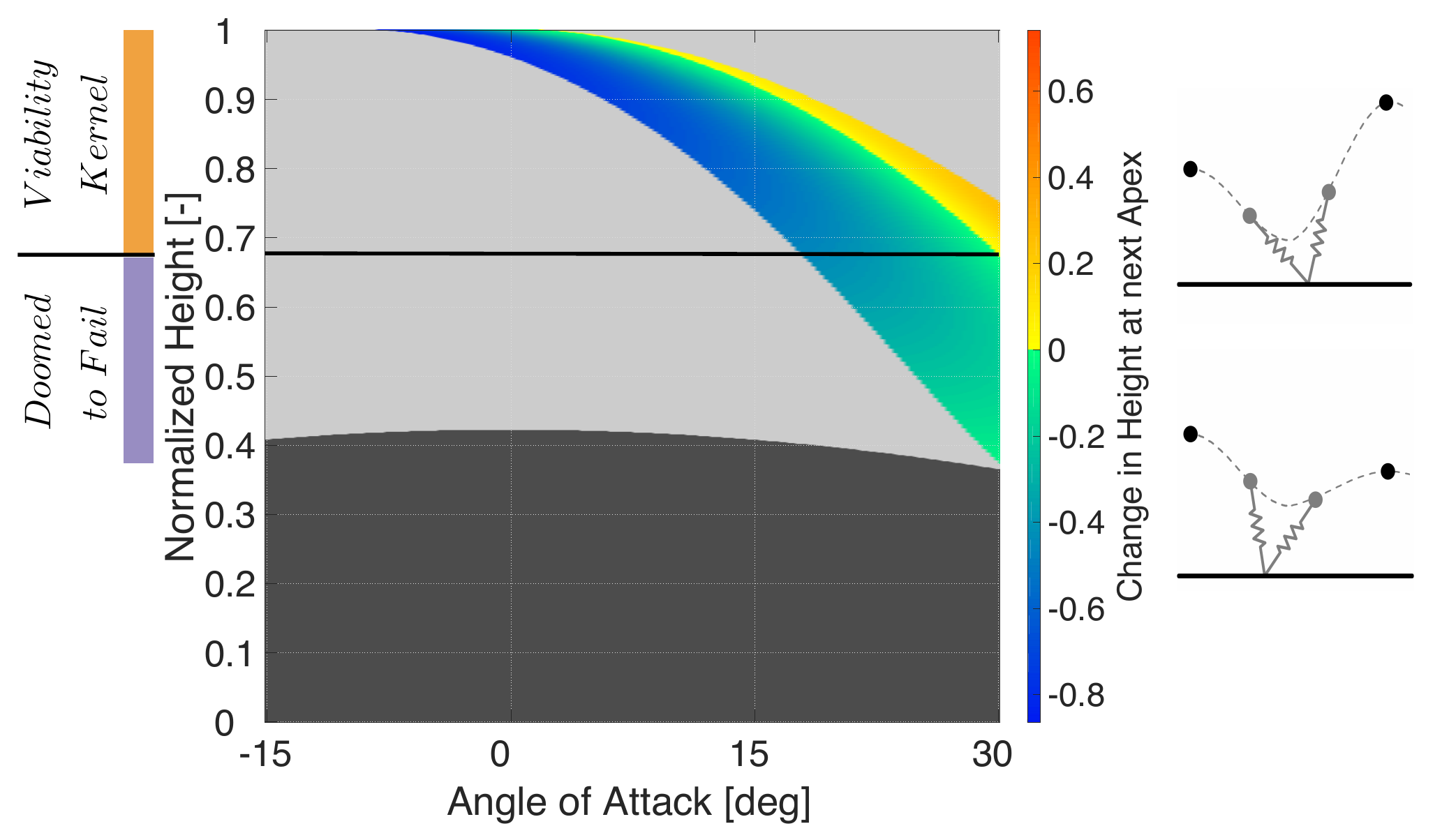}
    \caption{Represented is the change in state for the SLIP model, with an energy content of $E_g + E_k = 1.86 kJ$. The grey regions represent state-action pairs that result in falls, and the black regions are infeasible points: the foot would start underground at apex. The warm colored region result in a higher apex height, whereas the cold colored region results in a lower apex height. The black line at normalized height $0.675$ indicates the lower bound of the viability kernel: any states below this height can only select actions that lead to a lower hopping height, until it inevitably falls.}
    \label{fig:transmat}
\end{figure*}

Having reduced both state and action space to a single dimension, we obtain the entire transition matrix by brute force simulation of a finely discretized grid over $\bar{s}$ and $\alpha_k$, use this to compute a difference in state and visualize the result in Fig. \ref{fig:transmat}. 
The grey regions are failures ending with the body hitting the ground, and the colored regions are all the state-action combinations that result in a second hop. The warm colored region represent actions that lead to a higher height at the next apex, whereas the cold-colored region are actions that result in a lower height at the next apex. The intersection between these two regions are limit-cycles where the apex height remains constant. As the end-goal of locomotion is arguably to reach a specific distance, actually traveling on a limit-cycle is not strictly necessary: a policy which hops erratically can also reach the end-goal. The black region represents state-action combinations that are infeasible: the foot would start underground at apex. \par
This constraint between \ang{0} and \ang{30} for $\alpha_k$ is convenient to illustrate our case, while also corresponding to the usual range of angles used in human running \cite{seyfarth2002movement}. It is also a realistic constraint a robot might need to cope with, for example due to mechanical hard-stops, limitations in hip-swing velocity or to remain within friction cones. \par
The viability kernel of this constrained SLIP-model can be seen by inspecting Fig. \ref{fig:transmat}, and is the set of normalized apex heights $\bar{s}_k \in \left [0.675,\ 1\right]$. For each state in this set, actions can be chosen that either keep the system on a limit-cycle, bring it to a higher, or to a lower apex height. Thus, it is always possible to stay inside the set, making it a viability kernel \cite{aubin2009viability}. Any state that starts below the normalized height threshold of $0.675$, marked with a horizontal line in Fig. \ref{fig:transmat}, can only choose actions that either immediately fall or at best hop at subsequently lower apex heights until it falls. These states are \emph{doomed to fail}.

\section{Control and Learning}
In our study, our goal is to learn a control policy for choosing $\alpha_k$ at each apex which travels as far as possible from any viable initial state without falling. \par

\subsection{Control Policy}
\begin{figure}[tb]
    \centering
    \medskip Linear Gaussian Policies\par\medskip
    \includegraphics[width=0.95\columnwidth]{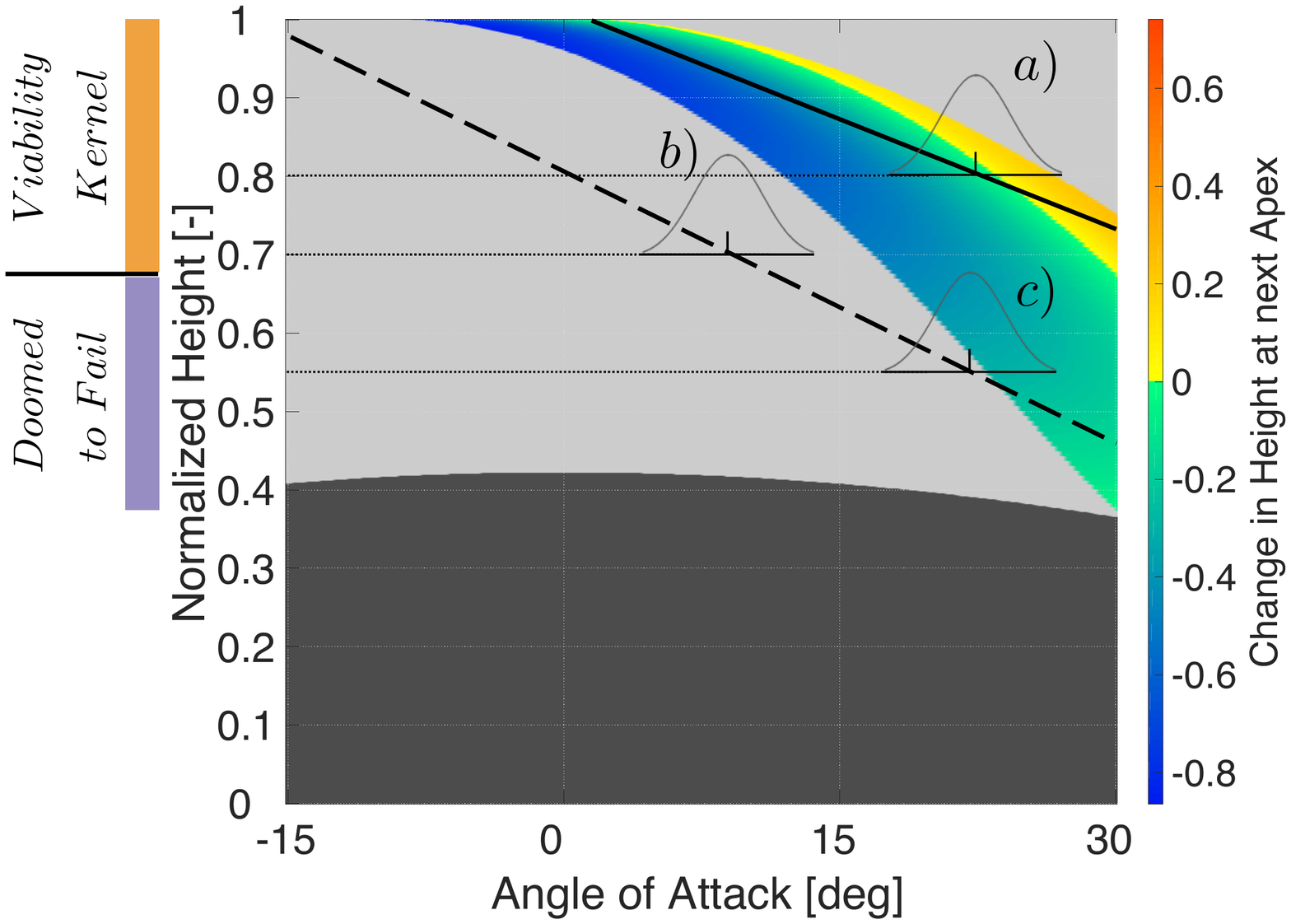}
    \caption{Represented are two linear controllers. The solid line is a well-tuned controller that would effectively stabilize the system, whereas dashed line is a guess that has the correct general shape, but is far from the salient gradient set (SGS). To highlight the problem, we have overlaid the gaussian sampling for three possible situations: in $a)$ the gaussian policy would sample with high probability from the SGS for a state inside the viability kernel. With the second greedy controller in $b)$ however, the probability of sampling a non-failing action is close to 0. Case $c)$ shows that the second greedy controller has a decent chance of sampling successfully from the SGS when initialized outside the viability kernel. It can learn from these samplings even though all initializations are doomed to fail.} 
    \label{fig:controllers}
\end{figure}

Various control policies have been proposed for SLIP models, often emphasizing deadbeat control \cite{wu20133}\cite{palmer2014periodic} or some form of optimality \cite{cnops2015basin}\cite{piovan2013two}. At its simplest, even a linear controller is able to stabilize any of the limit cycles, even if not in an optimal manner. We limit this study to a linear controller, visualized in Fig. \ref{fig:controllers}, since its low dimensionality allows us to directly visualize the reward landscape in parameter space, as in Fig. \ref{fig:landscapes}. The core of our results depends on the structure of the transition matrix (the dynamics) both inside and outside the viability kernel, which is independent of the parameterization used. \par
Specifically, we use a linear gaussian policy
\begin{align*}
\alpha_k \sim \mathcal{N}(\mu,\,\sigma^{2}) \\
\mu = \theta_0\, s_k + \theta_1
\end{align*}
where $\mathcal{N}$ is the normal distribution, $\mu$ is the mean and represents the greedy policy in absence of exploration, and $\sigma^2$ is the variance (our exploration parameter). The parameters $\theta_0$ and $\theta_1$ are the slope and offset of our linear policy\footnote{Though technically affine, we use the common practice of calling this linear to avoid confusion.}. Gaussian policies effectively include exploration directly into the policy and are often effective in continuous state and action systems, as often encountered in robotics \cite{sutton1998reinforcement}\cite{kober2013reinforcement}. 

\subsection{Interplay of Parameter Initialization, State Initialization and Exploration Strategy}
As an illustrative example, we award a fixed reward at each apex, which equates to learning to hop as many times as possible, regardless of the initial conditions. Using other periodic rewards, such as distance traveled per step, does not change the quality of the results. \par
It is clear that in order to learn, the agent needs to sample at least once a state-action pair that transitions to another apex: if the agent immediately samples from the state-action pairs that result directly in a fall, colored grey in Fig. \ref{fig:transmat}, it will have a constant reward of 0 and no gradient to learn from. For convenience, we will call the set of state-action pairs that result in a second apex height a \emph{salient gradient set (SGS)}. \par
With a random initialization of the policy parameters $\theta_0$ and $\theta_1$, it is possible that the greedy policy lies completely outside the SGS of the transition matrix, and therefore has a low chance of sampling from the SGS. 
The standard solution of increasing the variance of the gaussian policy has a drawback. While it increases the probability of sampling from the SGS when the greedy policy is initialized far away from the SGS, it also increases the probability to sample failing actions when the greedy policy is well tuned, as visualized with the solid line in Fig. \ref{fig:controllers}. \par
In this work, we highlight another aspect of the problem which, to the best of our knowledge, has been largely overlooked. In fig. \ref{fig:controllers} we show an example where the greedy policy chooses an action far from the salient gradient set for any state belonging to the viability kernel. In these cases, even local exploration will have only a low probability of ever taking a second step and receiving a reward. The same policy has a high probability of sampling non-failing actions for states \emph{outside} the viability kernel. In other words, we can directly change the reward landscape by including these states in the state initialization.

\subsection{Landscapes}


\begin{figure*}[tbh]
  \centering
  \medskip
  Reward Landscapes\par\medskip
  \begin{subfigure}[b]{0.45\textwidth}
  \centering
  \text{\small Landscapes with Viable Initialization}\par\medskip
    \includegraphics[width=\textwidth]{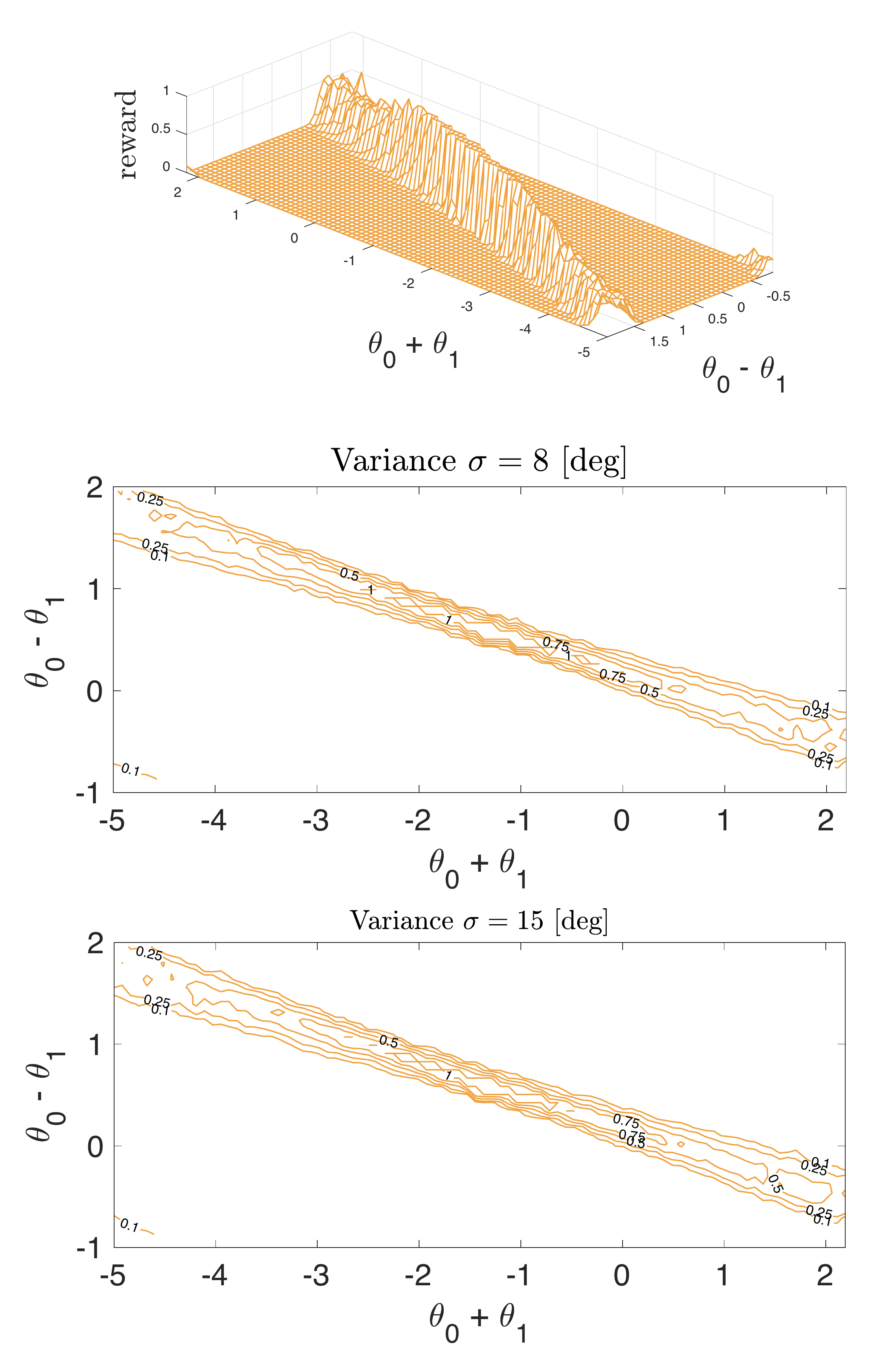}
  \end{subfigure}
  \begin{subfigure}[b]{0.45\textwidth}
  \centering
  \text{\small Landscapes with Feasible Initialization}\par\medskip
    \includegraphics[width=\textwidth]{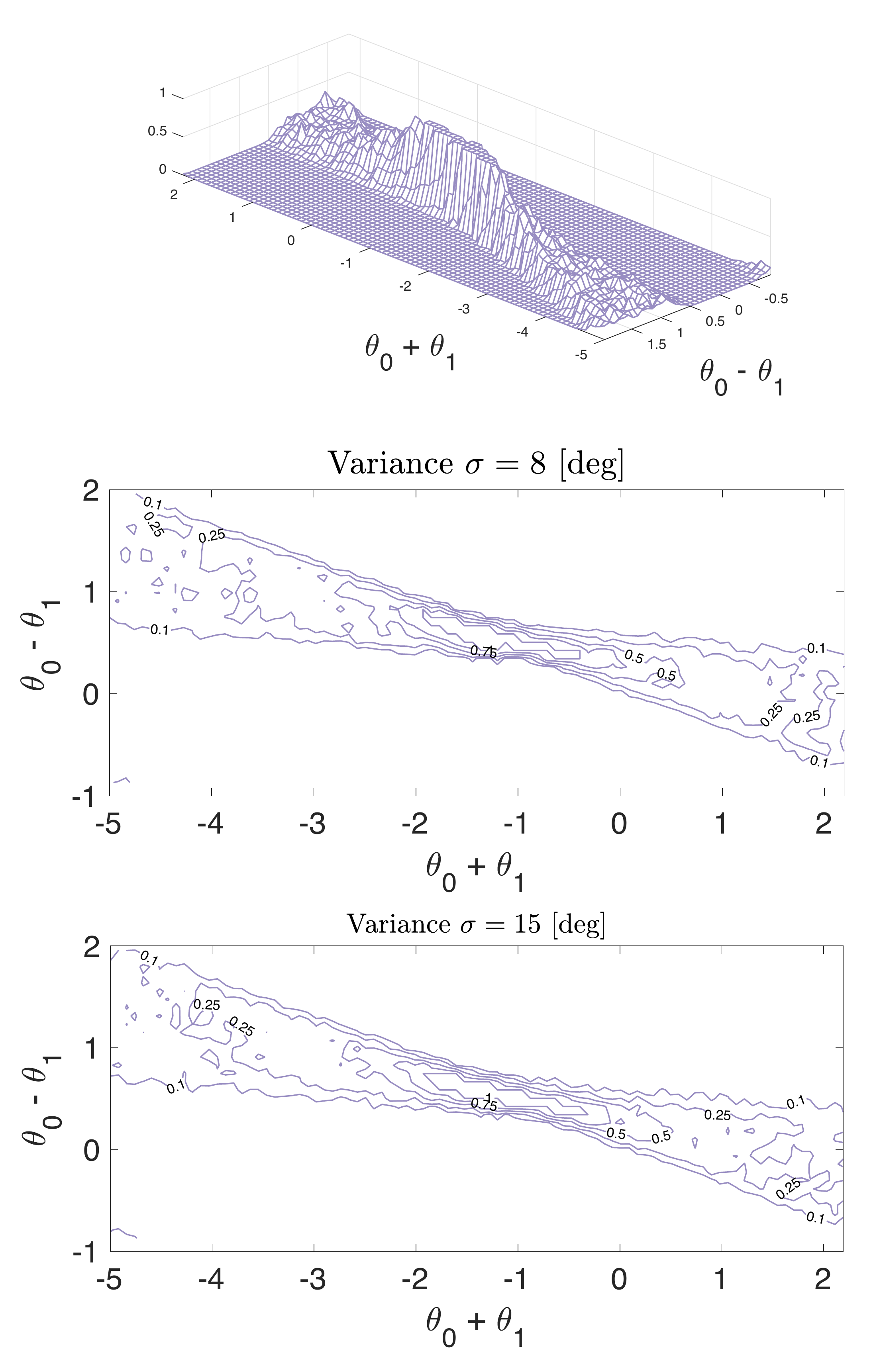}
  \end{subfigure}
  \caption{In the reward landscapes on the left, the system is initialized with a uniform random sampling inside the viability kernel, whereas on the right it is initialized from states with a feasible action, whether viable or doomed to fail. For visual clarity, the parameter space has been rotated 45 degrees and the reward (number of steps taken) is capped at 1. At values greater than this, the gradient is very steep and policies quickly become stable for many steps. The meshed landscapes (top row) are shown for clarity and are the same as the first set of contours (second row) with variance of \ang{8}. The landscape on the right has a salient gradient set (a point in parameter space with non-zero gradient) just over 79\% larger. The landscapes with a more aggressive variance of \ang{15} (bottom row) are shown to highlight how little effect increasing exploration has in comparison to the choice in state initialization.}
  \label{fig:landscapes}
\end{figure*}
We discretized over the parameters $\theta_0$ and $\theta_1$, computed 100 roll-outs for each point of the grid using a fixed exploration rate $\sigma$ and used the average return as an estimate of the reward. We then interpolated over the grid to obtain the reward landscapes. \par
We repeated this for two scenarios: strictly \emph{viable initializaiton} (left column in Fig. \ref{fig:landscapes}) and \emph{feasible initializations} (right column in Fig. \ref{fig:landscapes}. Viable initializations are initial states sampled with a uniform distribution from the viability kernel, whereas feasible initializations are sampled from a uniform distribution of all feasible states, including states that are doomed to fail. \par
We have also repeated this estimate for different variance, to highlight the much greater effect the state initialization has on the reward landscape. Indeed we found that using feasible initialization increased the SGS by just over 79\%. Nearly doubling the variance, from \ang{8} to \ang{15}, only increased the SGS by 13\%. This highlights a case where a large effort to find effective exploration strategies can be replaced with an improved state initialization.


\subsection{Learning Setup and Results}
To test our results, we implemented a standard temporal-difference algorithm with eligibility trace, a typical policy-gradient class of reinforcement learning algorithms \cite{sutton1998reinforcement}\cite{williams1992simple} relying on the likelihood ratio \cite{sutton2000policy} to estimate the policy gradient. To find appropriate hyperparameters for the learning step size and discount factor, several hundred learning trials were run with randomized hyperparameters, which revealed most consistent performance with a very low discount factor around $0.95$, and learning step sizes on the order of magnitude of $0.001$; in the presented results, these values are used. The variance of the gaussian policy begins at \ang{15} and is reduced to \ang{10}, \ang{7.5} and \ang{5} as the average performance reaches 2, 3 and 4 steps respectively. Once a policy averages at least 15 steps, learning is terminated. \par
To compare learning performance we ran learning trials starting from the same initial policy parameter, using first only viable initializations and then feasible initializations. In the second case, once the policy averages more than three steps, we consider the policy to have learned sufficiently and switch to a viable initialization, since the end-goal is to be stable for viable initializations. \par
\begin{figure}[thb]
    \centering
    \medskip Rate of Success for Randomized Initial Policies \par\medskip
    \includegraphics[width=1\columnwidth]{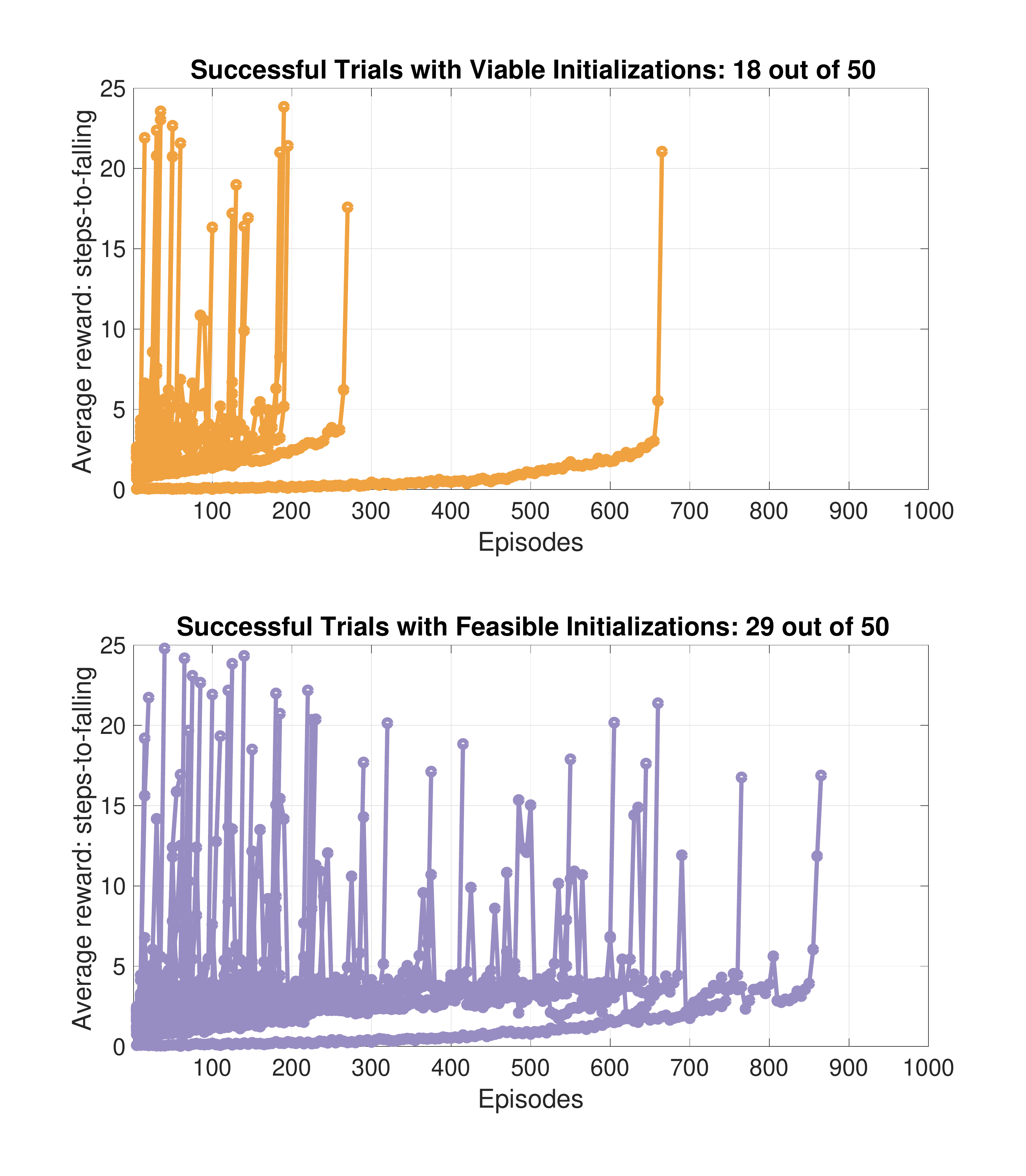}
    \caption{Fifty initial policy parameters were tested, sampled at random with a minimum chance of taking a 0.5 steps on average when using a viable initialization. For visual clarity, only successful trials are plotted.}
    \label{fig:random50}
\end{figure}

We first ran trials for 50 randomized policy initializations (see Fig. \ref{fig:random50}).
Before accepting each randomly selected policy initialization, its performance is estimated by averaging the reward over 100 roll-outs with viable initialization, as we did when estimating the reward landscape. Policy parameters with less than $0.5$ steps on average are discarded and re-sampled. In other words, we biased initializations towards policy parameters for which viable initializations also had a chance to learn, instead of including policy parameters where viable initializations had no chance of learning. \par
\begin{figure}[hbt]
    \centering
    \medskip Trial Policy Parameters\par\medskip
    \includegraphics[width=1\columnwidth]{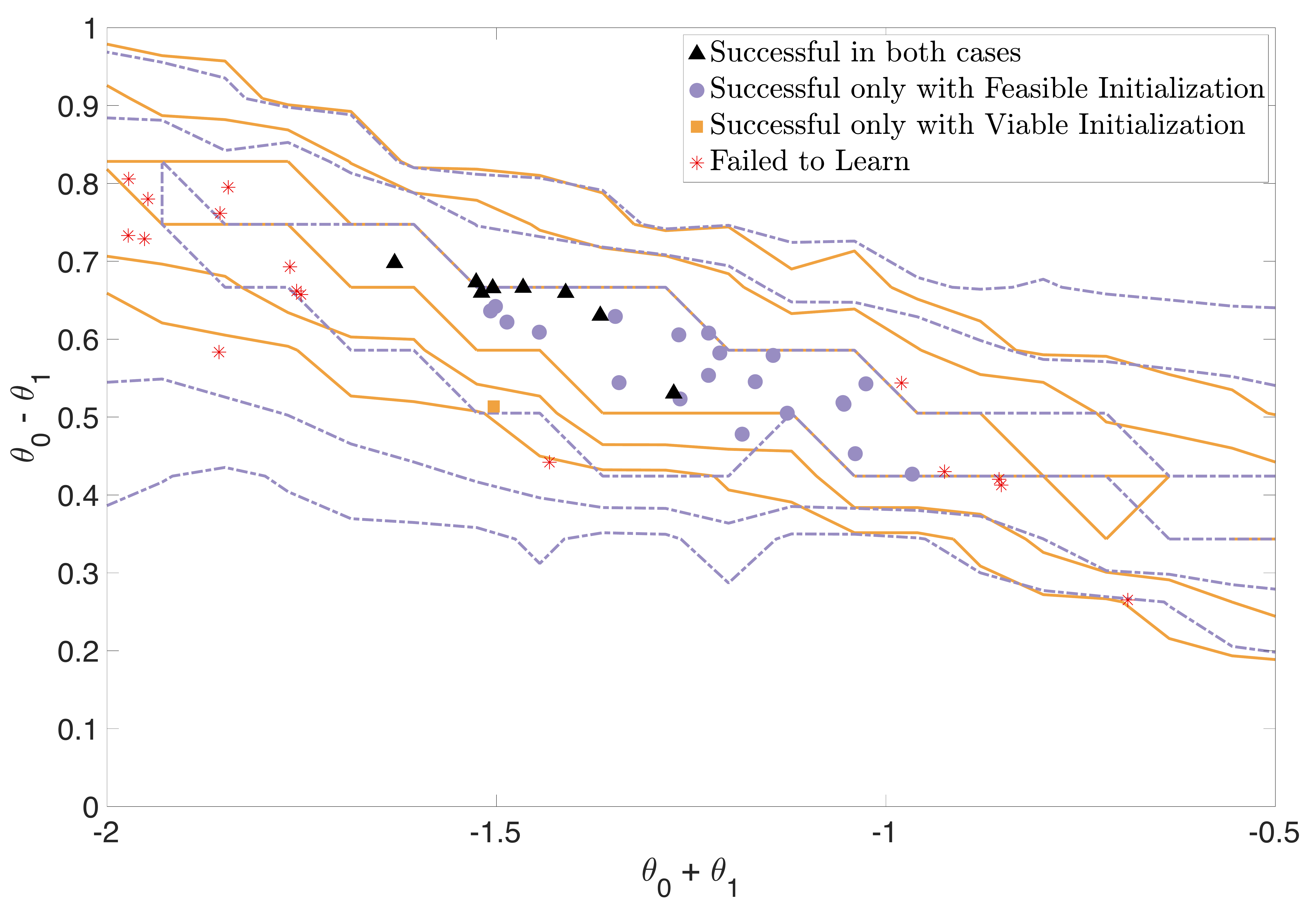}
    \caption{Examining the location of the initial policy parameters on the reward landscape shows just how brittle the viable initializations are. The distribution of the parameters which fail for both initializations also suggests some further structure we have not explored here. The landscape contours with viable initializations (solid orange line) and feasible initializations (dashed purple line) are overlaid for reference.}
    \label{fig:paronplots}
\end{figure}
\begin{figure}[htb]
    \centering
    \medskip Rate of Success for Multiple Trials\par\medskip
    \includegraphics[width=1\columnwidth]{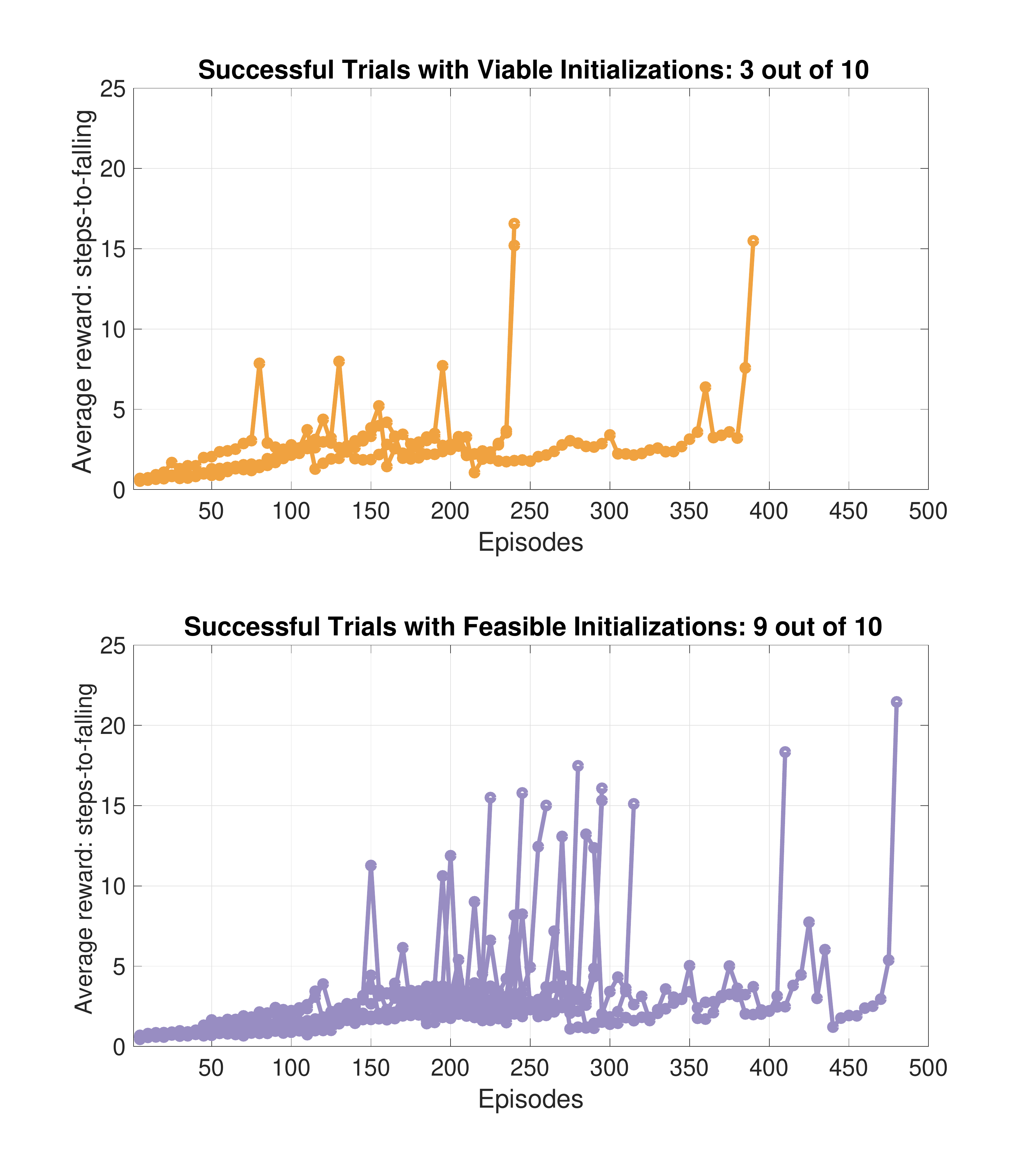}
    \caption{Ten trials starting from the same initial policy parameters were repeated, with viable initializations (top) and feasible initializations (bottom). The improved reliability with feasible initializations is clear, with 9 out of 10 trials succeeding, compared to 3 out of 10 for the viable initializations. For visual clarity, only successful trials are plotted.}
    \label{fig:single}
\end{figure}

As shown in Fig. \ref{fig:random50}, we found that starting with feasible initializations resulted in 60\% more trials ending in success. The policy initializations relative to the landscapes is shown in Fig. \ref{fig:paronplots}.
We then selected a single parameter at random, and repeated 10 trials each with viable and feasible initializations, shown in Fig. \ref{fig:single}. 
Again, we found much greater reliability in learning success when starting from feasible initialization rather than strictly viable initializations. 


\section{Discussion and Outlook}

We show the effect state initialization has on the reward landscape of a dynamically unstable system, and a clear case where a naive policy initialization learns more reliably when unviable states, that is states that are doomed to fail, are included in the initializations. We have in particular highlighted it is to consider the effect of state initialization in overcoming a common problem in robot learning: reward landscapes with large patches devoid of gradient information. This adds another tool to design learning setups, in addition to more established approaches of exploration strategies or reward shaping. \par
One of the major drawbacks of initializing in unviable states is that failures often result in damage to the robot. A solution is to start learning in simulation before transferring the policies to real robots \cite{peng2017sim}\cite{yu2017preparing}. With this approach, it is important not to mirror the same conditions of the robot in simulation, but purposefully explore with more aggressive and potentially unviable initializations. \par
While this approach is promising, a simulation still relies on a model even if the policy is learned model-free. One of the attractions of model-free learning is the reduced need for expert knowledge and engineering effort in order to directly deploy robots in the field. To this end, our results suggest a different emphasis for robot design is needed. A robot should not only be mechanically sturdy so it can survive failures, but there should be meaningful actions to be explored in unviable states. As we show with the SLIP model (see Fig. \ref{fig:transmat}), the choice to fall immediately or to stumble several steps before falling is what allows the system to learn from unviable states. This is a property of the system dynamics, and therefore the hardware design and not the controller design. Especially when building legged robots that try to mimic a SLIP-like behavior \cite{hubicki2016atrias}\cite{lakatos2017eigenmodes}, these aspects should be considered in addition to measures such as passive stability and energy efficiency.

\printbibliography

\end{document}